\pdfoutput=1
\documentclass[10pt,twocolumn,letterpaper]{article}
\usepackage{cvpr}
\usepackage{times}
\usepackage{epsfig}
\usepackage{graphicx}
\usepackage{multirow}
\usepackage{amsmath}
\usepackage{amssymb}
\usepackage{authblk}

\cvprfinalcopy % *** Uncomment this line for the final submission

\def\cvprPaperID{13} % *** Enter the CVPR Paper ID here

\begin{document}

%%%%%%%%% TITLE
\title{\ TAL EmotioNet Challenge 2020 \\
Rethinking the Model Chosen Problem in Multi-Task Learning}
\author{Pengcheng Wang, Zihao Wang, Zhilong Ji, Xiao Liu, Songfan Yang and Zhongqin Wu \authorcr
TAL Education Group \authorcr
{\tt\small \{wangpengcheng2, wangzihao3, jizhilong, liuxiao15, yangsongfan, wuzhongqin\}@100tal.com}}
\maketitle
%\thispagestyle{empty}

%%%%%%%%% ABSTRACT
\begin{abstract}
This paper introduces our approach to the EmotioNet Challenge 2020. We pose the AU recognition problem as a multi-task learning problem, where the non-rigid facial muscle motion (mainly the first 17 AUs) and the rigid head motion (the last 6 AUs) are modeled separately. The co-occurrence of the expression features and the head pose features are explored. We observe that different AUs converge at various speed. By choosing the optimal checkpoint for each AU, the recognition results are improved. We are able to obtain a final score of 0.746 in validation set and 0.7306 in the test set of the challenge.
\end{abstract}

%-------------------------------------------------------------------------
%%%%%%%%% BODY TEXT
\section{Introduction}
Facial Action Unit (AU) recognition is a difficult task, because AUs are defined by local and subtle changes in facial expressions. 
Early researchers use handcraft-features to recognize AU, but those manual shallow features are usually not discriminative enough for capturing the facial morphology \cite{zhao2015joint}. 
With the development of deep learning, neural networks have been increasingly used in facial representation learning for more effective AU recognition and detection \cite{zhao2016deep}. 
Existing deep learning methods usually require enough training data with precise labels. 
However, the EmotioNet challenge, one of the largest AU dataset, only provides 25k images with precise labels, which may result in over-fitting. 
Inspired by the spirit of multi-view co-regularization semi-supervised learning \cite{niu2019multi}, we train two deep neural networks to generate multi-view features. A multi-view loss is used to enforce the features to be conditionally independent. And a co-regularization loss is designed to make the predictions of the two views to be consistent.
We further find that the multi-view loss and the co-regularization loss also benefit supervised training, and the result can be better than semi-supervised training. So in this challenge, we train the multi-view co-regularization method as our base model, and then use the base model to filter the noisy data. 

Another important topic of AU recognition is how to model the co-occurrence relationship between correlated AUs.
Kaili \etal \cite{zhao2015joint} point that there exist strong probabilistic dependencies between different AUs. \emph{E.g.}\ AU1 (inner-brow raise) has a positive correlation with AU2 (outerbrow raise) and a negative correlation with AU6 (cheek raiser). 
In this challenge, we categorize the 23 AUs into two groups. 
The first group contains the first 17 AUs, and the second group contains the last 6 AUs. 
We use this division because AUs in the second group are about head pose recognition problem, which is different from the other emotion recognition problem (the first 17 AUs). 
Intuitively, AUs in different group should not have strong probabilistic dependencies. 
The data augmentation strategies of the two groups are also different. 
For instance, face alignment is important for the first group but is improper for the second group. 
Hence two separate models for the two groups are then trained to exploit dependencies among co-occurring AUs and facial features. 
The training of each model can be treated as a multi-task learning problem. 
We then find an interesting phenomenon that different AUs have diverse convergence speed. By choosing the best checkpoint for each AU, the final score can be obviously boosted.

The contents of this report are organized as follows. Section 2 presents the details of our method. In section 3, we present some experimental results. Section 4 concludes the report .

%------------------------------------------------------------------------
\section{Methods}

In this section, we first introduce the data utilization in our experiment.
We then show how we use the multi-view co-regularization method as our base framework. 
We finally analyze the multi-task model-chosen problem and present what strategy we use in detail. 

%-------------------------------------------------------------------------
\subsection{Data utilization}

As one of the largest AU dataset, EmotioNet 2020 contains 23 classes of AU in various scenarios. 
It includes 950k images with noisy labels, 25k optimization images with precise labels, 100k validation images and 200k testing images. 
Participants of the challenge can obtain the noisy images and optimization images. Because the noisy images are automatically labeled by machine and the accuracy is only 81\% \cite{benitez2017emotionet}, we cannot directly use them to train models. 

We randomly divide the optimization images into dataset-A (22.5k) for training and dataset-B (2.5k) for validating in our experiments. 
The head pose recognition problem (the last 6 AUs) is different from the other emotion recognition problem (the first 17 AUs), because it requires more context of the face area.

For the first model, we use RetinaFace \cite{deng2019retinaface} to detect and align the face, and conduct online data augmentation including random down sampling, random flipping, and random black square occluding as mentioned in \cite{benitez2017emotionet}.
For the second model, we use RetinaFace to detect faces (the face box's length is expanded to 1.5 times the original one) and conduct online data augmentation including down sampling and black square occluding.
We follow \cite{benitez2017emotionet} to expand dataset-B by 4 times in order to fit the distribution of the official testing dataset. 

To utilize the noisy data, we first use our base model to predict the label for each noisy image.
We then choose to use the label only if it is the same as the official given noisy label and the prediction probability is larger than a pre-defined threshold. 
After filtering the noisy images, we obtain dataset-C that contains 300k images. 
We use dataset-A and dataset-C to obtain a pre-trained model and finally fine-tune the model using dataset-A.

%-------------------------------------------------------------------------
\subsection{Multi-view Co-regularization AU Recognition}

Recently, inspired by the idea of co-training, many semi-supervised training methods are used to recognize AU. 
We follow the multi-view co-regularization method \cite{niu2019multi} as our base framework, but the difference is that we only use labeled data.
Instead of semi-supervised learning, we use the co-training idea as a constraint of our supervised training. The purpose is to capture multi-view features to enhance the model discrimination.

%---------------------update--------------
We use two backbone models (two efficientnet-b4 \cite{tan2019efficientnet} or two resnet-101\cite{he2016deep})
%----------------------------------
pre-trained on ImageNet as feature extractors. For each image, we denote $f_{i}$ as the i-th view feature that generated by the i-th feature extractor. Then two classifiers are trained to predict the probabilities of the j-th AU using $f_1$ and $f_2$:
$$p_{ij}=\sigma(w^{T}_{ij}f_{i}+b_{ij}),$$
where $\sigma$ denotes the sigmoid function, $w_{ij}^{T}$ and $b_{ij}$ are the respective classifier parameters.

As mentioned in Niu \etal  \cite{niu2019multi}, $f_1$ and $f_2$ are supposed to be conditional independent multi-view features, so a multi-view loss is used to orthogonalize the weights of the AU classifiers of different views. The multi-view loss $L_{mv}$ is defined as:
$$L_{mv}=\frac{1}{C}\sum_{j=1}^{C}\frac{W^{T}_{1j}W_{2j}}{\left\|W_{1j}\right\|\left\|W_{2j}\right\|}$$
where $W_{ij}=[w_{ij}, b_{ij}]$ denotes the parameters of the j-th AU’s classifier of the i-th view.

The classifiers of different views should get consistent predictions by minimizing the Jensen-Shannon divergence between the two predicted probability distributions \cite{niu2019multi}, and the co-regularization loss is defined as
$$L_{cr}=\frac{1}{C}\sum_{j=1}^{C}(H(\frac{p_{1j}+p_{2j}}{2})-\frac{H(p_{1j})+H(p_{2j})}{2}),$$
where $H(p)=-(p\log p+(1-p)\log(1-p))$.

For the final output, we use the average of two view's output:
$$p=\frac{(p_{1}+p_{2})}{2}.$$

We thus use four losses to train the models: recognition losses of two views, the multi-view loss and the co-regularization loss. 

%----------------------------------------------------------------
\subsection{Multi-task learning}
Multi-label can be treated as a special multi-task problem, 
where each task predicts a 0/1 label. 
There are many successful attempts of multi-task learning \cite{deng2019retinaface}.

%-----------figure-1---------------------
\begin{figure}[t]
\centering
\includegraphics[height=6.5cm,width=9.0cm]{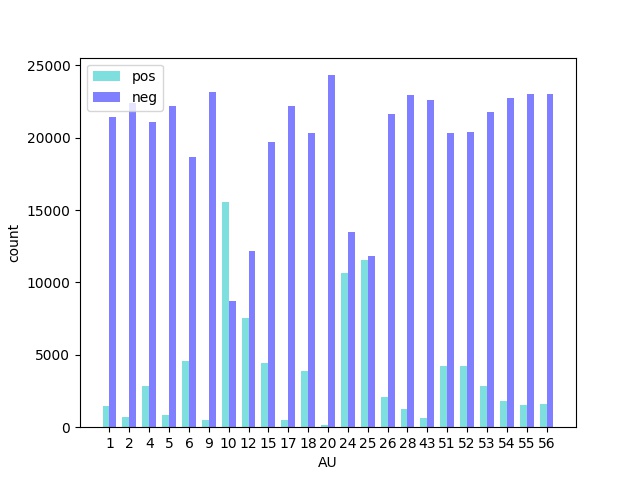}
\caption{The positive and negative numbers of each AU in the optimization dataset.}
\end{figure}
%---------------------------------------
% -----------fugure-2---------------------
\begin{figure}[h]
\centering
\includegraphics[height=6.5cm,width=8.0cm]{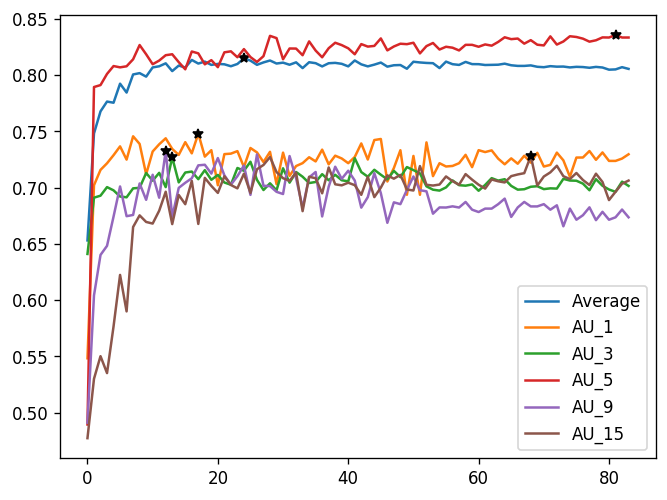}
\caption{Average, AU1, AU3, AU5, AU9, AU15 final scores changes along with epochs. The black star means the best final score of each line. As we can see, the best average final score is around the 23-th epoch, and the AU5 best final score is around the 80-th epoch. We can conclude that with the training process going, some AU may have been over-fitted, while others are still under-fitted.}
\end{figure}
% --------------------------------------

%------------------table-1-------------------
\begin{table*}[t]
\centering
\label{tab:my-table}
\begin{tabular}{|l|l|l|l|l|}
\hline
 &                                                   & F1     & Accuracy & Final score \\ \hline 
\multirow{6}{*}{17AU}       & Baseline                                       & 0.6740 & 0.9330   & 0.8030      \\ \cline{2-5} 
                      & Ensemble                                       & 0.6913 & 0.9435   & 0.8174      \\ \cline{2-5} 
                      & Ensemble-multicrop                             & 0.6985 & 0.9461   & 0.8223      \\ \cline{2-5} 
                      & Ensemble-multicrop-pretrained                  & 0.7033 & 0.9486   & 0.8259      \\ \cline{2-5} 
                      & Ensemble-multicrop-pretrained-chosen           & 0.7232 & 0.9490   & 0.8361      \\ \cline{2-5} 
                      & Ensemble-multicrop-pretrained-chosen-threshold & 0.7393 & 0.9514   & 0.8454      \\ \hline
\multirow{4}{*}{6AU}        & Baseline                                       & 0.3628 & 0.8708   & 0.6168      \\ \cline{2-5} 
                      & Ensemble                                       & 0.3916 & 0.8574   & 0.6245      \\ \cline{2-5} 
                      & Ensemble-chosen                                & 0.4068 & 0.8633   & 0.6350      \\ \cline{2-5} 
                      & Ensemble-chosen-threshold                      & 0.4148 & 0.8644   & 0.6396      \\ \hline
\end{tabular}
\vspace{5pt}
\caption{The results of 17 AUs and 6 AUs that trained on dataset-A and evaluated on dataset-B. ``Baseline'' means the multi-view framework. ``Ensemble'' means combining the results of  several models with different backbone by using linear model. ``Multicrop'' means using multi-crop strategy during inference. ``Pretrained'' means the models are pretrained on dataset-A + dataset-C and finetuned on dataset-A. ``Chosen'' means choosing different best checkpoint for each AU individually. ``Threshold'' means using different threshold to judge the label.}
\end{table*}
% --------------------------------------

We believe that there are relations between different AUs \cite{zhao2015joint}. For instance, AU 6 and AU 12 are known co-occur in expressions of enjoyment and embarrassment. 
We categorize the 23 AUs into two groups. The 6 head pose AUs are in a group and the other AUs are put in another group. 
We then train the AUs in the same group together through multi-task learning. 

We analyze the label distribution of the optimization dataset and find that the positive and negative ratio of each AU can be very unbalanced, as shown in Figure 1.
In order to balance the positive and negative samples, batch balancing \cite{hand2018doing} is adopted.
In our experiment, a hyper-parameter $\alpha = 0.2$ is used to balance the data distribution. 
If $\alpha$ times the number of negative samples is larger than the number of positive samples, we randomly select a subset of negative samples to calculate loss and ignore the others. 
For example, if the batch size is 100, the number of negative sample is 90 and the number of positive sample is 10.  Because $0.2 \times 90 > 10$,  we randomly select 18 negative samples and ignore other negative samples.

As multiple AUs are trained by the same model, we guess that their convergence speed may be different. We show the final scores on dataset-B of different AUs during training in Figure 2, and find an interesting phenomenon that the convergence speed of different AU can be very diverse. With the training process going, some AU may have been over-fitted, while others are still under-fitted.
Based on the above observation, we choose the best checkpoint for each AU, and the final score can be obviously boosted than choosing the best checkpoint based on the average.
Furthermore, since each AU in the multi-task training has inconsistent convergence speed, the threshold for judging the AU label should also be different. 
We show the experimental results of model and threshold chosen in the next section.

\begin{table*}[t]
\centering

\begin{tabular}{|l|l|l|l|l|}
\hline
                                   & Group                   & Mean Accuracy & F1     & Final score \\ \hline
\multirow{3}{*}{The validation phrase}   & TAL                     & 0.9200        & 0.5720 & 0.7460      \\ \cline{2-5} 
                                   & University of Magdeburg & 0.9198        & 0.5706 & 0.7452      \\ \cline{2-5} 
                                   & SIAT-NTU                & 0.9195        & 0.3531 & 0.6363      \\ \hline
\multirow{4}{*}{The testing phrase}      & TAL                     & 0.9147        & 0.5465 & 0.7306      \\ \cline{2-5} 
                                   & University of Magdeburg & 0.9124        & 0.5478 & 0.7301      \\ \cline{2-5} 
                                   & SIAT-NTU                & 0.9013        & 0.4410 & 0.6711      \\ \cline{2-5} 
                                   & USTC-alibaba            & 0.8609        & 0.3497 & 0.6053      \\ \hline
\end{tabular}
\vspace{5pt}
\caption{The results of the validation phrase and the testing phrase. In both phrases, we rank the 1st among the teams}
\end{table*}
%------------------------------------------

%---------------------tabel-2-----------------------
\begin{table}[]
\begin{tabular}{|l|l|l|l|}
\hline
                    & F1     & Accuracy & Final score \\ \hline
w/o $L_{mv}$             & 0.6447 & 0.9592   & 0.8019      \\ \hline
w/o $L_{cr}$             & 0.6188 & 0.9583   & 0.7885      \\ \hline
w/o batch balancing & 0.6446 & 0.9593   & 0.8019      \\ \hline
Baseline         & 0.6556 & 0.9604   & 0.8080      \\ \hline
\end{tabular}
\vspace{5pt}
\caption{The results of 12 AUs (defined by the EmotioNet challenge 2018). Models are trained on dataset-A and evaluated on dataset-B. When we remove $L_{mv}$, $L_{cr}$, batch balancing, the final score descends 0.0061, 0.0195, 0.0061 respectively.}
\end{table}

\section{Experiments}
The evaluation criteria of the challenge is the average of $F_{1}$ score and mean accuracy: 
$$Final \: score=\frac{accuracy+F_{1}}{2}.$$

We summarize the results of 17 AUs and 6 AUs on dataset-B with different strategies in Table 1. 
As to the result of 17 AUs, the base model obtains a final score of 0.8030 and the ensemble of multiple base models obtain a final score of 0.8174.
By adding the multi crop strategy, the final score increases 0.0049.
By adding the pre-training strategy, the final score increases 0.0036.
By adding the model chosen strategy, the final score increases 0.0102.
By adding the threshold strategy, the final score increases 0.0093.
As to the result of 6 AUs, we tried some similar strategies, but only model chosen and threshold strategy are useful. 
The performance gains are 0.0105 and 0.0046 respectively.

%------------------------------------------

%-------------table-3---------------------

In the official validation and testing datasets, we combine all the strategies motioned above and obtain the final scores of 0.7460 and 0.7306, both place the 1st among the teams. The results are  presented in Table 2.

In Table 3, we compare the results of our base model and the model that without multi-view loss, co-regularization loss, or batch balancing. 
Removing the co-regularization loss $L_{cr}$ causes the largest performance gap, which verifies the results that reported in \cite{niu2019multi}. 

\section{Conclusion}

In our submission to the EmotioNet challenge 2020, we use the multi-view co-regularization framework as our baseline. 
By adding checkpoints and threshold chosen strategies, we boost the performance by a large margin and rank the 1st in the challenge. A more thorough and systematic research will be done in our future work. 
%---------------------------------------------------

% {\small
\bibliographystyle{ieee_fullname}
\bibliography{egbib}
% }

\end{document}